# Supervised Image Segmentation for High Dynamic Range Imaging

Ali Reza Omrani[1,2] [a], Davide Moroni[1] [b]
[1] *Institute of Information Science and Technologies (ISTI), National Research Council of Italy, Pisa, Italy*
[2] *Department of Engineering, Università Campus Bio-Medico di Roma, Rome, Italy*
*{ali.omrani, davide.moroni}@isti.cnr.it*

Abstract: Regular cameras and cell phones are able to capture limited luminosity. Thus, in terms of quality, most of the produced images from such devices are not similar to the real world. They are overly dark or too bright, and the details are not perfectly visible. Various methods, which fall under the name of High Dynamic Range (HDR) Imaging, can be utilised to cope with this problem. Their objective is to produce an image with more details. However, unfortunately, most methods for generating an HDR image from Multi-Exposure images only concentrate on how to combine different exposures and do not have any focus on choosing the best details of each image. Therefore, it is strived in this research to extract the most visible areas of each image with the help of image segmentation. Two methods of producing the Ground Truth were considered, as manual threshold and Otsu threshold, and a neural network will be used to train segment these areas. Finally, it will be shown that the neural network is able to segment the visible parts of pictures acceptably.

## 1 INTRODUCTION

Natural scenes have a vast luminosity; however, regular cameras are capable of capturing a limited dynamic range of that luminance. Therefore, the generated image has regions with High- (overly bright) and Low- Exposure (too dark), and the detail is not visible well. These types of pictures are called Low Dynamic Range (LDR) images.

The first solution to cope with this problem is to utilise cameras with special sensors, which can obtain more luminance than regular cameras and are able to produce images with more details and more similar to the real-world (Nayar & Mitsunaga, 2000; Tumblin & et al., 2005; McGuire & et al., 2007; Tocci & et al., 2011; Hajisharif & et al., 2015; Zhao & et al., 2015; Serrano & et al., 2016). However, due to the high cost of such equipment, it is not affordable and usable for regular users.

Another solution for this issue is using software development methods, which are known as High Dynamic Range (HDR) imaging. Various algorithms have been proposed recently, and the existing techniques can be divided into HDR imaging with Single-Exposure and Multi-Exposure methods. In the Single-Exposure algorithms, various techniques can produce an HDR image starting from a single LDR image. However, these methods are not satisfying since the detail cannot be restored goodly. In (Eilertsen & et al., 2017), the authors proposed an algorithm to generate an HDR image from an LDR image. Still, their method was affected by two problems: the inability to reconstruct details of dark and overly saturated areas. More precisely, this algorithm was not able to retrieve the details in the excessively saturated regions. Therefore, in (Omrani & et al., 2020) proposed to first merge input images with different exposures and afterward fed the wavelet coefficient of the merged image to the network to produce more details in a shorter time.

Fortunately, unlike the Single-Exposure methods, Multi-Exposure ones are more effective and can reconstruct more detail. More clearly, several LDR images are combined in such techniques and produce an HDR image. Although Multi-Exposure methods perform almost perfectly on static scenes, they can encounter problems such as ghosting in dynamic scenes due to moving objects. However, several algorithms have been proposed to solve this problem (Green Rosh & et al., 2019; Kalantari & Ramamoorthi, 2017; Wu & et al., 2018; Yan & et al., 2019; Prabhakar & et al., 2019; Prabhakar & et al., 2020).

---

[a]  https://orcid.org/0000-0001-6506-1830
[b]  https://orcid.org/0000-0002-5175-5126

Additionally, deep learning has been a great help in the scope of computer vision in recent years. For instance, (Eilertsen & et al., 2017) used a deep neural network to produce an HDR image in the logarithmic domain. Also, (An & Lee, 2017) used deep learning to reconstruct the detail of an image with different row-wise exposure in the irradiance domain. The works (Lee & et al., 2018; Endo & et al., 2017), unlike other methods, used neural networks to produce several LDR images with different exposures from a single LDR image. Additionally, (Kalantari & Ramamoorthi, 2017) first aligned images with the optical flow method and eventually used deep learning to fuse the aligned images to produce an image with more details. Moreover, in (Green Rosh & et al., 2019), two deep learning methods were used to align images and generate an HDR image. Furthermore, (Yan & et al., 2019) used neural networks with different scales of images to learn the relative relation between input images and their Ground Truth.

Image Segmentation is one of the tasks in computer vision whose objective is to simplify image analysis. This task is typically used to detect objects or better understand images, such as medical ones. Also, image segmentation can be utilised to extract the regions of images with more details. In (Vadivel & et al., 2004), the authors analysed images in HSV colour space to segment pixels based on the value of Intensity or Hue. Additionally, other works proposed two methods for image segmentation based on luminance: histogram division (Kinoshita & Kiya, 2019) and clustering based on the Gaussian Mixture Models (GMM) of the histogram (Kinoshita & Kiya, 2018). Furthermore, (Lee & Sunwoo, 2021) proposed a method to find the optimal valley point based on the slope between the histogram value of each pixel and other neighbouring points and used that valley point to segment regions.

Our Contributions are as follows:
- We will propose two methods to extract the best areas of images with more details.
- We will compare the proposed methods to specify the best one.

## 2 PROPOSED METHOD

### 2.1 Producing Ground Truth

Most proposed algorithms in the realm of HDR imaging are concentrated on how to produce them, while less attention has been paid to extracting suitable features. Thus, the proposed method in this research focuses on how to extract the most suitable regions for HDR imaging. Because by finding the areas with more details, the HDR algorithm can produce an image free of overly saturated or dark parts. More specifically, an Image segmentation method is proposed to segment areas with the most detail. Therefore, a neural network can be utilised to extract the desired regions of input images, which will be discussed in a future work. Additionally, two different methods, i.e. manual thresholding and Otsu segmentation, were used to produce the Ground Truth, which will be compared with each other.

As mentioned before, two methods were used to produce Ground Truth. In the manual technique, several experts investigated the best possible range of intensity in YCbCr color space for extracting the areas with the most detail empirically. Eventually, an average of the scopes was calculated for each image. The selected ranges for image intensity with Low and High-Exposure are [120,255] and [0,200], respectively. Generally, the objective is to acquire areas with less darkness and saturation. Therefore, because most of the regions in Low-Exposure images are dark, we would like to extract the areas with the highest pixel values, which indicate the most visible ones.

Conversely, because most pixels in High-Exposure images are saturated, the objective is to extract pixels with the lowest values. Certainly, by choosing pixel values in the luminance channel, some of the visible pixels with the lowest values cannot be selected. For example, although the grey area of the mountain in Fig. 1 is visible, it was not selected in the segmentation process.

The second method is called the Otsu technique, which calculates a threshold based on the intensities of images and segment pixels. More precisely, the pixels greater than the threshold are considered foreground (white), and those with lower values as background (black). The difference between these two methods is that the Otsu technique threshold is computed based on the histogram of each image. Whereas in the manual, all the pictures of each exposure have the same range. Moreover, in Otsu, all the pixels of Low-Exposure images greater than the threshold are considered the desired pixels, while the pixels lower than the threshold in High-Exposure pictures are desirable.

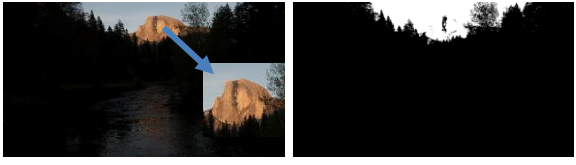

Figure 1. The image on the left is the input image, and on the right is its Ground Truth produced by the manual method. The picture is taken from (Fairchild, 2007).

## 2.2 Neural network Structure

Unfortunately, each image has various intensities, and it would be challenging to use non-machine learning methods to predict them. Moreover, it is a time-consuming task to extract a range for each image separately. Therefore, a neural network has been proposed in this research to learn how to extract the best area of each image based on the proposed ranges in the training stage.

Two U-Net-shaped networks were used for segmentation in this research, and each network is trying to learn how to map from each exposure to its Ground Truth. As can be seen in Fig. 2, the U-Net consists of 2 parts. In the first part, the subnetwork strives to extract features, and the second subnetwork tries to produce an output similar to the Ground Truth. The encoder section includes five blocks, and each block has two convolutional layers with ReLU function, DropOut, and MaxPool layers, respectively. Additionally, kernels of convolutional layers in each block are 16,32,64,128,256, respectively. Moreover, the decoder has four blocks, and each block consists of one transpose convolutional, concatenate, convolutional layer with ReLU activation, Drop Out, and another convolutional layer with ReLU, respectively. Furthermore, all convolutional and transpose convolutional layers used a kernel size of 3x3, and the last layer used a kernel size of 1x1.

## 2.3 Loss Functions

The loss function is one of the essential components in deep learning. Thus, to select the best loss function for segmenting the regions with the most detail, three loss functions will be used and compared. The used loss functions are as follows:
1. Binary Cross Entropy (BCE): One of the most common functions, which is used in most image segmentation research is the BCE

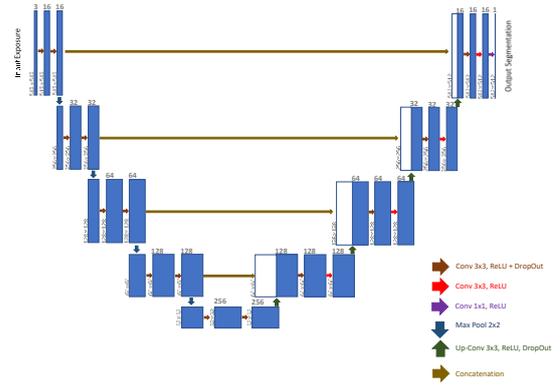

Figure 2. Total scheme U-net Architecture, which was used in this experiment. The blue boxes denote feature maps; their number is on the top of each box, and their size is indicated on the lower left side of each box.

loss function, and it can be represented as follows:

$$L_{BCE} = -\sum (y \log \hat{y} + (1-y) \log(1-\hat{y})) \quad (1)$$

Where y and $\hat{y}$ represent Ground Truth and the network's output, respectively, and the sum is over all the pixels.

2. Focal Loss: This loss function is used for imbalance data and focuses on hard data:

$$L_{focal} = -\sum (\alpha \cdot y \cdot (1-\hat{y})^\gamma \log(\hat{y}) + (1-\alpha)(1-y)(\hat{y})^\gamma \log(1-\hat{y})) \quad (2)$$

Where α and γ are hyper-parameters and, as a default, they are equal to 0.25 and 2.0, respectively.

3. Combo Loss (Dice Cross-Entropy): This loss function is also used for imbalanced data and is produced by a combination of Cross-Entropy and Dice loss functions. Eq (3) represents Dice loss, and Eq (4) is for Combo loss:

$$DL(y, \hat{y}) = 1 - \frac{2y\hat{y} + 1}{y + \hat{y} + 1} \quad (3)$$

The number one added to the numerator and the denominator avoids undefined errors, such as y=$\hat{y}$=0.

$$L_{DiceCE} = L_{Dice} + L_{BCE} \quad (4)$$

## 3 EXPERIMENT RESULTS

### 3.1 Dataset

Recently, a new dataset was collected for High Dynamic Range (HDR) Imaging Challenge called NTIRE 2021 (Perez-Pellitero & et al., 2021). In this dataset, two types of pictures (Single-Exposure and

Multi-Exposure images) were provided for this challenge; however, Multi-Exposure images only were used in this research. More specifically, this dataset includes images from (Froehlich, & et al., 2014) that were generated as follows. First, HDR images were produced natively by two Alexa Arri cameras with a mirror rig; then, their corresponding LDR images were generated synthetically with noise sources. There are approximately 1500 pairs of HDR/LDR images in this dataset for the training set, 40 for the validation set, and 200 pictures for the test set with a resolution of 1900x1060. Moreover, all the images were already aligned and saved after gamma correction.

## 3.2 Evaluation Metrics

Several evaluation parameters have been used in this research to evaluate the results and will be discussed as follows:
1. Dice Index: This metric is region based and evaluates the similarity and the overlaps of two samples.
$$Dice\ (A, B) = 2\ \frac{|A \cap B|}{|A| + |B|} \qquad (5)$$
2. Jaccard Index: This metric works similarly to Dice and calculates the similarity of two samples.
$$Jaccard\ (A, B) = \frac{|A \cap B|}{|A \cup B|} \qquad (6)$$
3. Two other metrics are Sensitivity and Specificity, which calculate True Positive and True Negative pixels.
$$Sensitivity = \frac{TP}{TP + FN} \qquad (7)$$
$$Specificity = \frac{TN}{TN + FP} \qquad (8)$$
4. Area under Curve: Another evaluation metric is AUC which is commonly used in image segmentation algorithms.
$$AUC = 1 - \frac{1}{2}\left(\frac{FP}{FP + TN} + \frac{FN}{FN + TP}\right) \qquad (9)$$

## 3.3 Ground Truth Generation

As the used dataset is not consisting of Ground Truths for segmentation, the first objective of this research is to produce Ground Truths that cover the most area of scenes. Thus, after frequent and visual studying of produced Ground Truths by both manual and Otsu techniques, it became evident that the manual method has more coverage than the latter one. For instance, as can be seen in Fig. 3, both approaches worked almost the same on images with Low-Exposure. However, the manual method succeeded in covering more areas in images with High-Exposure. Additionally, as can be seen on the last row, the total covered area by the manual method is more than the Otsu technique. Therefore, the produced Ground Truth from the manual method will be used for the rest of the research.

## 3.4 Other Details

Additionally, the training process for each loss function was 50 epochs, which took around 200 minutes on NVIDIA DGX Station A100 and less than two minutes for testing, and the model was trained as parallel on 4 GPUs. Moreover, the number of images for the training set and the validation set was about 1300 and 200 images with a resolution of 512x512 and a batch size of 32, respectively. Furthermore, Adam optimiser with a learning rate of 0.001 was used. Finally, the neural network was implemented in Tensorflow (Keras) framework.

Moreover, during experiments, three input images with different exposures were used for image segmentation, in which, after obtaining the suitable areas of Low- and High-Exposure images, the remaining regions were extracted from the Medium-Exposure images. However, the acquired areas of the Medium-Exposure were not sensible because most of them were only a few pixels that didn't have any shapes. Thus, it was difficult for the network to segment them. Fig. 4 demonstrates an example of the extracted regions in the Medium-Exposure image.

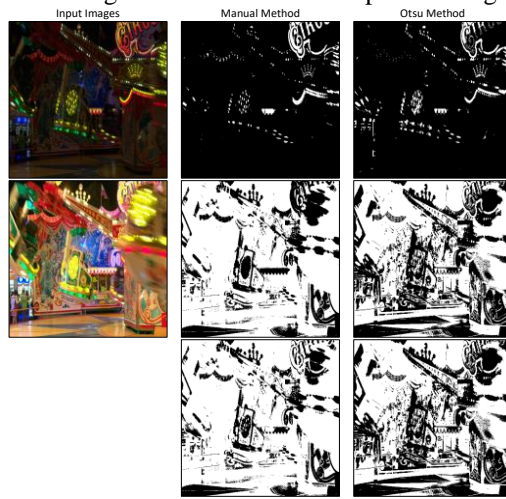

Figure 3. Produced Ground Truth of both Manual and Otsu Methods. First row is generated from the Low-Exposure image, the second one is obtained from the High-Exposure image, and the third row is a merged output of both rows.

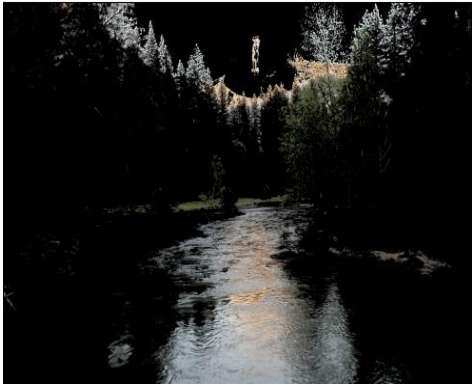

Figure 4. An example of extracted areas from a Medium-Exposure image. The picture is taken from (Fairchild, 2007).

Table 1. Quantitative evaluation results of Low-Exposure Image Segmentation. M row determines metrics, which are specified as M1: Dice, M2: Jaccard, M3: Sensitivity, M4: Specificity, M5: AUC, and AVG is the average of the metrics.

| Loss functions | M1 | M2 | M3 | M4 | M5 | AVG |
|---|---|---|---|---|---|---|
| BCE | 0.951 | 0.905 | 0.912 | **0.999** | 0.498 | 0.853 |
| Focal | 0.916 | **0.936** | **0.997** | 0.997 | 0.498 | **0.869** |
| Dice - BCE | **0.965** | 0.933 | 0.912 | **0.999** | 0.498 | 0.861 |

Table 2. Quantitative evaluation results of High-Exposure Image Segmentation. M row determines metrics, which are specified as M1: Dice, M2: Jaccard, M3: Sensitivity, M4: Specificity, M5: AUC, and AVG the is average of the metrics.

| Loss functions | M1 | M2 | M3 | M4 | M5 | AVG |
|---|---|---|---|---|---|---|
| BCE | **0.994** | 0.909 | 0.765 | 0.754 | **0.68** | **0.82** |
| Focal | 0.989 | 0.89 | 0.753 | **0.763** | 0.675 | 0.814 |
| Dice - BCE | 0.991 | **0.912** | **0.77** | 0.73 | 0.67 | 0.815 |

## 3.5 Results

The predicted segmentation outputs by three different loss functions were compared quantitatively with their produced Ground Truth by manual technique. As can be seen in Table 1, which demonstrates the evaluation results of Low-Exposure Image Segmentation, different loss functions outperformed the others in different evaluation metrics. For instance, the Focal Loss function performed better than the other two in Jaccard and Sensitivity evaluation metrics. Although they have equal values in the AUC evaluation metric, the Focal loss averagely was better than Dice-BCE and BCE. Additionally, Table 2 indicates that the Dice-BCE loss function worked better than the other two losses in Jaccard and Sensitivity evaluation metrics, but as a result, BCE was better on average. Therefore, it can be concluded that Focal loss function is able to segment better illumination in Low-Exposures and BCE in High-Exposures.

Figs. 5 and 6 demonstrate produced outcomes by different loss functions for both images with Low- and High- Exposure. As can be seen, although all the outputs are almost identical visually and are difficult to distinguish differences between them, the quantitative results demonstrated that the output of Dice-BCE is not as well as the output of the other two. Moreover, Fig. 7 indicates more examples of losses.

## 4 FUTURE WORKS

As discussed in the proposed method section, Otsu and manual methods were used in this research, and in the manual technique, a range was computed empirically. Although experiments demonstrated that the empirical approach had better outcomes than Otsu, it has two cons. Firstly, failure to recognise dark visible areas, such as the mountain peak illustrated in Fig. 1. Secondly, if the segmentation process is performed with the manual technique, the calculated range needs to be applied to all images, and it is possible that the computed span is not suitable for some photos, and calculating a specific span for each picture is also a time-consuming task. Therefore, it is better to work on a new automatic technique to estimate these ranges for each image.

In addition to working on a novel method for calculating an automatic range for each image in future work, it is feasible to use extracted regions from segmentation techniques in HDR imaging to produce an HDR image with more details. Additionally, this work can help reduce the complexity of networks for generating an HDR image.

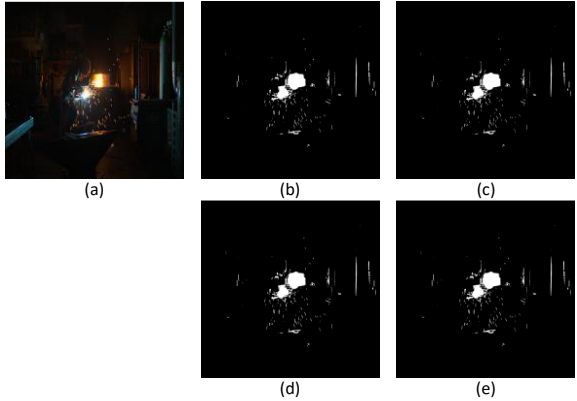

Figure 5. Output results of other losses. (a) Low-Exposure input image, (b) Dice-BCE Output, (c) BCE Output, (d) Focal Output, (e) Ground Truth.

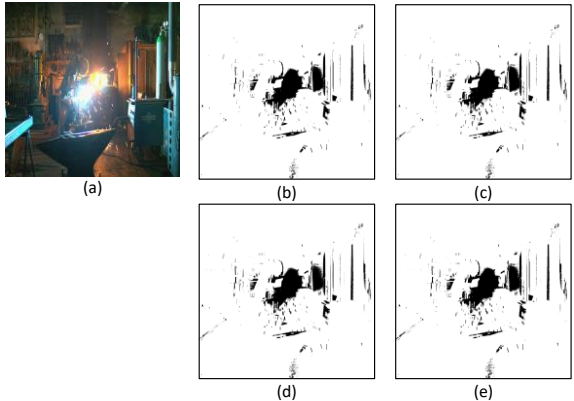

Figure 6. Output results of different losses. (a) High-Exposure input image, (b) Dice-BCE Output, (c) BCE Output, (d) Focal Output, (e) Ground Truth.

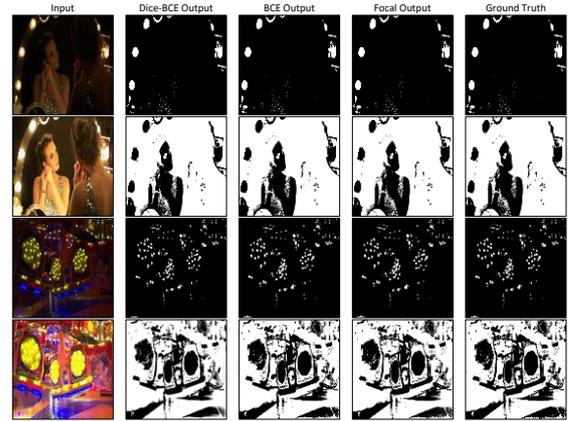

Figure 7. More examples with outputs.

## 5 CONCLUSION

Two methods for segmenting visible regions were used in this research, and a manual technique that is an empirical approach was chosen after comparing them to produce the Ground Truth. Moreover, deep neural networks were used to learn to extract the regions with the help of produced Ground Truths in each exposure. Additionally, three different loss functions were utilised in this article, and the quantitative metrics demonstrated that the focal and BCE loss functions outperformed in Low-Exposure and High-Exposure images, respectively.